\documentclass[review]{elsarticle}

\usepackage{lineno,hyperref}
\modulolinenumbers[5]

\usepackage{multirow}

\begin{document}

\begin{frontmatter}

\title{Classification supporting COVID-19 diagnostics based on patient survey data}


\author[polsl_ksisk]{Joanna Henzel\corref{contrib}}

\author[polsl_aed]{Joanna Tobiasz\corref{contrib}}

\author[polsl_ksisk]{Micha{\l} Kozielski\corref{mycorrespondingauthor}}

\author[polsl_AI]{Ma{\l}gorzata Bach}
\author[polsl_graph]{Pawe{\l} Foszner}
\author[polsl_ksisk]{Aleksandra Gruca}
\author[polsl_graph]{Mateusz Kania}
\author[polsl_aed]{Justyna Mika}
\author[polsl_aed]{Anna Papiez}
\author[polsl_AI]{Aleksandra Werner}
\author[polsl_aed]{Joanna Zyla}
\author[polsl_SUM]{and Jerzy Jaroszewicz}
\author[polsl_aed]{Joanna Polanska}
\author[polsl_ksisk]{Marek Sikora\corref{mycorrespondingauthor}}
\ead[url]{adaa.polsl.pl}

\cortext[contrib]{Those authors contributed equally to this paper, and should be regarded as co-first authors.}
\cortext[mycorrespondingauthor]{Corresponding author}
\ead{michal.kozielski@polsl.pl, marek.sikora@polsl.pl}

\address[polsl_ksisk]{Department of Computer Networks and Systems, Silesian University of Technology, Gliwice, Poland}
\address[polsl_aed]{Department of Data Science and Engineering, Silesian University of Technology, Gliwice, Poland}
\address[polsl_AI]{Department of Applied Informatics, Silesian University of Technology, Gliwice, Poland}
\address[polsl_graph]{Department of Graphics, Computer Vision and Digital Systems, Silesian University of Technology, Gliwice, Poland}
\address[polsl_SUM]{Department of Infectious Diseases and Hepatology, Medical University of Silesia, Katowice, Poland}

\begin{abstract}
Distinguishing COVID-19 from other flu-like illnesses can be difficult due to ambiguous symptoms and still an initial experience of doctors.
Whereas, it is crucial to filter out those sick patients who do not need to be tested for SARS-CoV-2 infection, especially in the event of the overwhelming increase in disease.
As a part of the presented research, logistic regression and XGBoost classifiers, that allow for effective screening of patients for COVID-19, were generated. 
Each of the methods was tuned to achieve an assumed acceptable threshold of negative predictive values during classification.
Additionally, an explanation of the obtained classification models was presented. The explanation enables the users to understand what was the basis of the decision made by the model.
The obtained classification models provided the basis for the DECODE service (decode.polsl.pl), which can serve as support in screening patients with COVID-19 disease. Moreover, the data set constituting the basis for the analyses performed is made available to the research community. This data set consisting of more than 3,000 examples is based on questionnaires collected at a hospital in Poland.
\end{abstract}


\end{frontmatter}


\section{Introduction}

The outbreak of the COVID-19 pandemic has created new challenges for the physicians around the world.
A new disease requires the development of new drugs, procedures and approaches to treating patients.
Additionally, diagnostics may be a challenge, especially when there are few obvious and unambiguous symptoms differentiating the disease from other infections.
The list of symptoms associated with COVID-19 does not allow for unequivocal distinction of COVID-19 from other influenza-like illnesses. Limited testing capacity is one of the biggest problems the healthcare systems are facing. Hence, it is crucial to filter out those sick patients who do not need to be tested for SARS-CoV-2 infection.

The diagnosis of the patient's condition requires experience supported by reliable and unambiguous straightforward diagnostic methods.
As long as rapid, inexpensive and reliable testing does not support the physicians in their work, experience is imperative.
Experience can be gained by analysing cases of subsequent patients arriving to the hospital. Besides, it can be enriched owing to a comprehensive analysis presenting conclusions for a larger set of data.

Literature studies show that machine learning methods were verified and gave good results for a number of diagnostic tasks \cite{caballe2020machine}. Various types of classification models were studied in the past for the diagnosis of patients with diseases such as for example Ebola~\cite{COLUBRI201954}, HIV~\cite{CHOCKANATHAN201924}, heart disease~\cite{GARATEESCAMILA2020100330}, cancer~\cite{SAXENA2020182} or diabetes~\cite{KAVAKIOTIS2017104}.
Referring to the listed applications of machine learning in diagnostics, it can be concluded that prognostic models can provide a valuable summary of clinical knowledge and can be useful when such expertise is unavailable \cite{COLUBRI201954}.

The choice of methods supporting patient classification and diagnosis may depend on the type of data analysed.
For example, diagnostics based on image analysis currently uses mainly deep learning methods, e.g. \cite{SAXENA2020182,Wynantsm1328}.
Other types of data, including data collected from questionnaires, are analysed using various methods, among which there can be identified a division (\cite{CHRISTODOULOU201912}) into statistical approaches represented by logistic regression and data-driven machine learning methods (e.g. decision tree based methods, SVM, Na\"ive Bayes, kNN). Many of the works, however, verify and compare the quality of several different models, e.g. \cite{GARATEESCAMILA2020100330,AHAMAD2020113661,ma_ng_xu_xu_qiu_liu_lyu_you_zhao_wang}.

The main motivation for this work was the interest of the medical community in what characterizes COVID-19 (e.g. as opposed to influenza) and whether it is possible to create classifiers of acceptable quality that can support diagnostics based on survey questions to identify people suffering from COVID-19.
Such solutions can be especially valuable for new diseases, when there is little experience available and when the number of patients is growing rapidly becoming overwhelming for the healthcare system, as in the case of COVID-19.

Another motivation was the need to learn and understand the COVID-19 disease from the possibly broadest perspective. As the research results show (\cite{Escobar17720}), the course of the disease for various reasons may be different for different populations and therefore the analysis of data from Poland may be valuable.

The study aims to provide the medical community, especially family doctors, with a tool to distinguish patients suffering from other common infections from individuals who are suspected for COVID-19 and need to be further diagnosed with more advanced molecular methods. The main intention was thus to limit the number of patients referred to genetic testing and still omit possibly few COVID-19(+) cases.

Therefore, the aim of this study is to verify two classification approaches as screening methods that can support the diagnosis of sick patients with COVID-19. The classification is intended to be based on the symptoms identified during the initial assessment of the patient condition. The methods utilised in the research are the classifiers that are the most popular and efficient in medical and tabular data analysis. 
Another goal of this work is to provide a new data set enriching the knowledge on the COVID-19 disease and to share the results of analysis concerning both the data and the generated classification models.

Contribution of this work is threefold.
Primarily it consists of generating the classifiers that are tuned to an assumed acceptable threshold of negative predictive values so that the results allow for effective screening of patients for COVID-19.
Besides, the contribution contains an explanation of the obtained classification models enabling the users to understand what was the basis of the decision made by the model.
Finally, in conjunction with this analysis, a new data set is shared, based on questionnaires collected in a hospital in Poland.
The last part of the contribution consists of the data set preparation, processing, characterization and visualization leading to the identification of COVID-19 characteristics. 
It is assumed that providing a data set along with its characterization and predictive analysis performed on this set will be a valuable opening for further analysis and meta-analysis.

The performed analysis resulted in an online service\footnote{https://decode.polsl.pl} available for anyone who needs support in COVID-19 diagnostics.

The structure of this paper is as follows.
Section \ref{sec:related_work} presents an overview of previous research related to the presented topics. Section \ref{sec:data} outlines the characteristics of the shared data set being the basis for classification models. Section \ref{sec:classification} presents data preparation steps and two applied approaches to classifier generation and tuning. Section \ref{sec:results} focuses on the evaluation of the models created. Section \ref{sec:discussion} presents the discussion and explanation of the results obtained. Section \ref{sec:app} outlines the developed web application that makes an initial diagnosis based on the given symptoms. Section \ref{sec:conclusions} concludes the paper.

\section{Related work}
\label{sec:related_work} 

Due to the great involvement of the scientific community in the research aimed at understanding the SARS-CoV-2 virus and the COVID-19 disease, many studies have recently been published dealing with this issue from different perspectives. Many works on the use of machine learning methods to diagnose COVID-19 have been covered in review articles such as \cite{SWAPNAREKHA2020109947}, \cite{Wynantsm1328} and \cite{LALMUANAWMA2020110059}.

The work \cite{SWAPNAREKHA2020109947} presents the analysis of the research dynamics in the field between January and May 2020.
In this work, based on the frequency of occurrence of various methods, the following classes of solutions were distinguished: Deep Learning approaches (CNN, LSTM and others), Mathematical and Statistical methods, Random Forest, SVM, and Others (e.g. Linear Regression, XGBoost). The described approaches were applied to various data types with a predominance of X-ray images and achieved good results. However, the lack of real data was explicitly highlighted in the work.  

An extensive meta-analysis (107 studies with 145 models) of the works published between January and April 2020 is presented in \cite{Wynantsm1328}. In this study, three classes of predictive models are distinguished, and these are models for use in the general population, for COVID-19 diagnosis and prognosis. Among 91 diagnostic models, 60 focus on image analysis, while 9 predict the disease severity. 
Out of the remaining 22 works presenting diagnostic models not based on imaging, only two studies were based on the data sets containing more than 1000 examples.

The review \cite{LALMUANAWMA2020110059} discusses among others machine learning in COVID-19 screening and treatment. The examples of approaches presented in this work are focused on image and clinical data (e.g. blood test results) based diagnostics.

Among the studies presenting diagnostic machine learning-based models \cite{Wynantsm1328}, the works presented below can be distinguished as the most interesting due to the size of a data set, questionnaire-based data features and utilised analytical approaches. 

The work \cite{Menni2020.04.05.20048421} presents a statistical analysis of data attributes and generation of multivariate logistic regression model on a data set consisting of 1702 individuals (579 were SARS-CoV-2 positive and 1123 negative), whose data were collected through the online application. The attributes describing each example included personal characteristics (sex, age, BMI) and flu-like symptoms (e.g. fever, persistent cough, loss of taste and smell, etc.).

The study \cite{Diaz-Quijano2020.04.05.20047944} presents an approach where based on routinely collected surveillance data a multiple model using logistic regression was generated. The data set that was the basis of the analysis consisted of 5739 patient records (1468 were SARS-CoV-2 positive and 4271 negative) collected in Brazil.

There are numerous data sets related to COVID-19 reported \cite{shuja2020covid}. Such data sets include information on COVID-19 diagnosis, case reporting, transmission estimation, sentiment analysis from social media and semantic analysis from the collection of scholarly articles.
Most of the COVID-19 diagnosis data are based on image (CT scans, X-ray images) analysis.

A data set [dataset]\cite{2020covidclinicaldata} that can be referred due to the subset of similar features was collected by Carbon Health, a healthcare provider in the US.
The anonymised data of Carbon Health patients were collected between March and June 2020. The data includes the clinical characteristics (epifactors, comorbidities, vitals, clinician-assessed symptoms, patient-reported symptoms) and laboratory results of patients on the date of service.

Another data set [dataset]\cite{kaggle_covid} was collected at the Hospital Israelita Albert Einstein, at Sao Paulo, Brazil. Hospital patients had samples collected to perform the SARS-CoV-2 RT-PCR and additional laboratory tests.
The created data set contains 5644 examples of which 10\% were SARS-CoV-2 positive.
The attributes consist of virus, blood and urinea test results and internal assignment to a hospital ward.
This data set was analysed in \cite{schwab2020predcovid} and its part was analysed in \cite{Batista2020.04.04.20052092}.
The study \cite{schwab2020predcovid} presents the results of several different classification methods (Logistic Regression, Neural Network (multi-layer perceptron), Random Forest, Support Vector Machine and Gradient Boosting) in terms of their quality. Three directions of analysis were chosen in this study covering identification of: patients that are likely to receive a positive SARS-CoV-2 test result, COVID-19 patients that are likely to require hospitalisation, and COVID-19 patients that are likely to require intensive care. Additionally, within each direction, the feature importance was identified.

In response to the growing demand for information on modern IT services developed in the healthcare sector, and in order to coordinate activities undertaken in this regard in Europe, the mHealth Innovation and Knowledge Hub was created on the initiative of WHO ITU/Andalusian Regional Ministry of Health \cite{European51:online}. It was established to collect and share experiences on modern e-medicine solutions and to support countries and regions in implementing large-scale activities in this regard.

 mHealth initiatives are especially important these days, when the World is facing SARS-CoV-2. Many governments, companies, and citizens movements have developed various mHealth solutions to keep the population informed and help manage the crisis situation. A repository of such solutions developed in Europe can be found in \cite{European43:online}. It is a dynamic resource that is updated as additional tools are reported.
 
 Among the many services created around the world, there can be found those that help to find out whether someone may have been exposed to the coronavirus in order to reduce the spread of the virus \cite{TheNHSCO40:online,ProteGO_Safe:online, Radar_COVID:online}. Others serve patients in a better understanding of the mechanisms of the COVID-19 or are the channels of updated information on regional regulations including territory-specific restrictions \cite{LocalCOV28:online, CERCACOVID:online}. Some of the solutions provide the most up-to-date research findings and information, including all the latest data on COVID-19 diagnosis and treatment thereby helping medical personnel make informed clinical decisions \cite{WHO_ACADEMY:online}. 
 
 There are also applications that, similarly to the service presented in this study, support the diagnosis of sick patients with SARS-CoV-2 infection based on the symptoms of the disease. For example, solutions presented in \cite{healthdi61:online, Testingf99:online, Koronawi16:online, COVID19R49:online, Mediktor72:online, TheHuman42:online, NuovoCor51:online, httpsasi60:online} allow to self-assess the possible symptoms of this infectious disease and to learn about the recommendations to be followed. Some of them give the estimated risk of SARS-CoV-2 infection as a result, however, they do not disclose the methods used in these estimates. In addition, some mentioned applications are available only in one language version (e.g. in Italian \cite{NuovoCor51:online}) or are intended for use in a given region or country and they require selecting the name of a specific locality/province when completing the questionnaire \cite{httpsasi60:online}. Other ones are not publicly available and need a special social security number assigned by local Health Service to activate the application \cite{Infosfur89:online}. Some, in turn, are targeted only for medical staff \cite{Hippokra35:online}.
 
 A relatively small group of applications \cite{intensivecare:online,herokuapp:online} are those developed as a result of published research \cite{Feng2020.03.19.20039099,Brinati2020.04.22.20075143}. The corresponding works concerned COVID-19 diagnostics using various machine learning methods. The common feature of the data analysed by them were attributes describing blood indices.

The state of the art survey presented above shows that there exist numerous approaches proving the usability of classification models in COVID-19 diagnostics. However, to the best of the authors' knowledge, there are no approaches reported as a research paper taking into consideration solely questionnaire data what can be valuable when the healthcare system is overloaded or clinical tests are not available. There are numerous online applications performing symptom-based diagnosis, however, no information on the applied methodology is given.

Another conclusion is that there are freely available data sets on the Internet, however the sets containing data of more than 1000 patients are scarce and it is still important to enrich the available data representation. Additionally, to the best of the authors' knowledge, there were no questionnaire-based data sets collected in Poland reported and it is worth filling this gap.

\section{Data set}
\label{sec:data}

The data set was collected at Specialised Hospital No. 1 in Bytom, Poland between 21\textsuperscript{st} February 2020 and 30\textsuperscript{th} September 2020 thanks to the cooperation of hospital staff and data scientists involved in the project. 

Data acquisition was a process starting at the hospital where people arrived in order to be diagnosed. Before any examination or testing, each person was asked to fill in the questionnaire containing questions relevant to COVID-19 diagnosis. Initially, the survey was containing few questions about the basic symptoms consisting of the occurrence of temperatures exceeding 38\textsuperscript{o}C, cough and dyspnoea. With time and increasing knowledge about the COVID-19 characteristics, the survey was enriched with further questions and options. When the final diagnosis resulting from SARS-CoV-2 test was known, the survey was anonymised, scanned, tagged with SARS-CoV-2 test result and sent to the database of survey images. Next, the information encapsulated in each survey was transformed into a feature vector, i.e. an example in our data set. 

The questionnaires were filled in by patients themselves. People waiting to be admitted to the hospital for further tests may feel unwell and anxious, which affects the consistency and quality of completing the questionnaires. Therefore, a synthetic and unambiguous form of the survey was developed, which is now utilised within the web application (see Fig. \ref{fig:questionnaire}) being one of the results of this research. Nevertheless, due to such manual nature of data collection, the data was subject to additional inspection in order to remove errors introduced while completing the survey.

The collected data set consisted of 3114 patient records. Each patient recorded in the collected data was described by 32 attributes. Within these attributes the following classes of patient characteristics can be identified:
\begin{itemize}
    \item 18 attributes describing symptoms,
    \item 7 attributes listing comorbidities,
    \item 3 attributes representing the patient's condition.
\end{itemize}
The other attributes include epidemiological attributes such as: age, sex, blood group and contact with infection.
Each patient was classified by two conditions:
\begin{itemize}
    \item Symptoms: Healthy / Sick
    \item SARS-CoV-2 test result: Positive / Negative
\end{itemize}
The quantitative characteristics of the collected sample from population on the basis of the classification listed above is presented in Table \ref{tab:data_char_quantity}. 

\begin{table}[!htb]
	\caption{Quantitative characteristics of the main analysed data set.}
	\label{tab:data_char_quantity}
	\centering
	\begin{tabular}{lccc}
    	\hline
	    &   SARS-CoV-2 (--)	&	SARS-CoV-2 (+)	&	Total	\\
	    \hline
	    Healthy	&	1000	&	173	&	1173	\\
	    Sick	&	1355	&	586	&	1941	\\
    	Total	&	2355	&	759	&	3114	\\
	    \hline
	\end{tabular}
\end{table}

In the presented study patients who experienced disease symptoms (Sick) were taken into consideration, i.e. 1941 people. From this group, 577 patients, collected mainly in August and September, were extracted to final classification evaluation (test set). The characteristic of test set taking into account SARS-CoV-2 diagnosis is presented in Table \ref{tab:data_char_quantity_add}
\begin{table}[!htb]
	\caption{Quantitative characteristics of the test data set.}
	\label{tab:data_char_quantity_add}
	\centering
	\begin{tabular}{lccc}
    	\hline
	    &   SARS-CoV-2 (--)	&	SARS-CoV-2 (+)	&	Total	\\
	    \hline
	    Sick	&	393	&	184	&	577	\\
	    \hline
	\end{tabular}
\end{table}

\subsection{Data preparation}
\label{sec:data_prep}

The first step of the initial data preparation was related to the aim of the conducted research, which was the screening of sick patients to help identify SARS-CoV-2 infections and, more importantly, to filter out only patients with other infections, who do not require molecular tests.
Having the defined goal in mind, the subset of patients identified as sick was selected from the collected data set presented in Table \ref{tab:data_char_quantity} and this subset was used in further analysis. 

\begin{figure}[!htb]
	\includegraphics[width=0.9\textwidth]{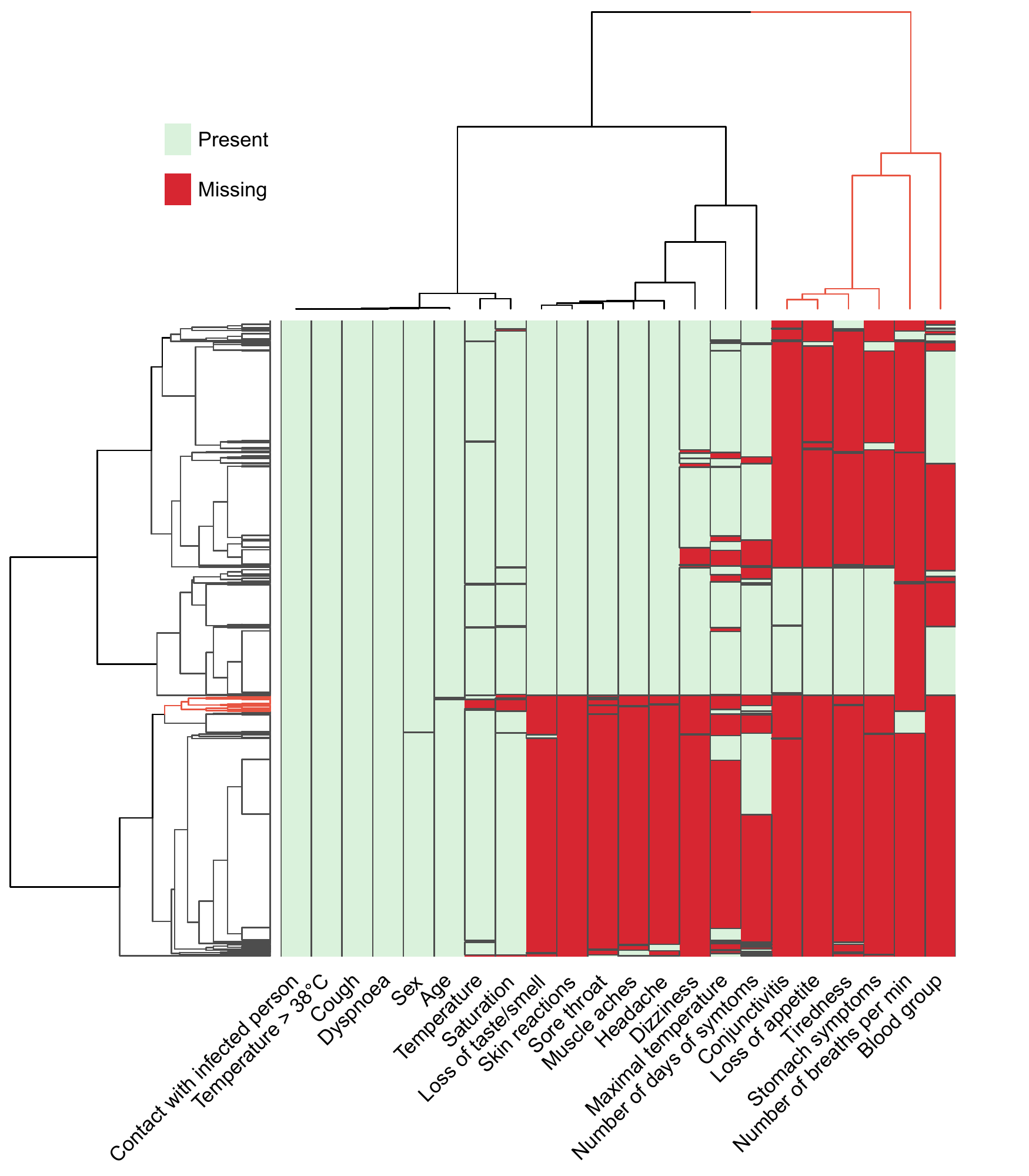}
	\caption{Feature and patient selection aimed at removing missing values on the basis of hierarchical clustering.}
	\label{fig:dendrograms}       
\end{figure}

Next, due to the fact that there were many missing values in the collected data, it was necessary to select the attributes and cases to generate the best possible classification model. In order to perform data selection, the data set was transformed into a binary representation where missing records were marked with the value of 1, while complete records with the value of 0. For the "contact with infected person" variable missing value was treated as a negative answer due to the survey construction: patients were asked to tick "Yes" option only if they knew they were exposed to SARS-CoV-2. Next, hierarchical clustering with the Hamming distance \cite{6772729} and McQuitty agglomeration \cite{doi:10.1177/001316446602600402} was performed for features and patients separately.
Finally, the cluster of patients with the highest level of missing information was removed from the data set and the cluster of features with the highest level of missing data was not considered in further analysis.
The dendrograms leading to the selection process and the data that was rejected in this selection are illustrated in Fig. \ref{fig:dendrograms}.
The data subjected to the selection are illustrated at the Fig. \ref{fig:dendrograms}, with missing and complete information marked with red and bright green colors, respectively. The dendrograms serving for the selection are at the sides of the figure. Branches highlighted in red correspond to patients or features excluded from the further analysis due to high missing data levels.

\begin{figure}[!htb]
	\includegraphics[width=0.9\textwidth]{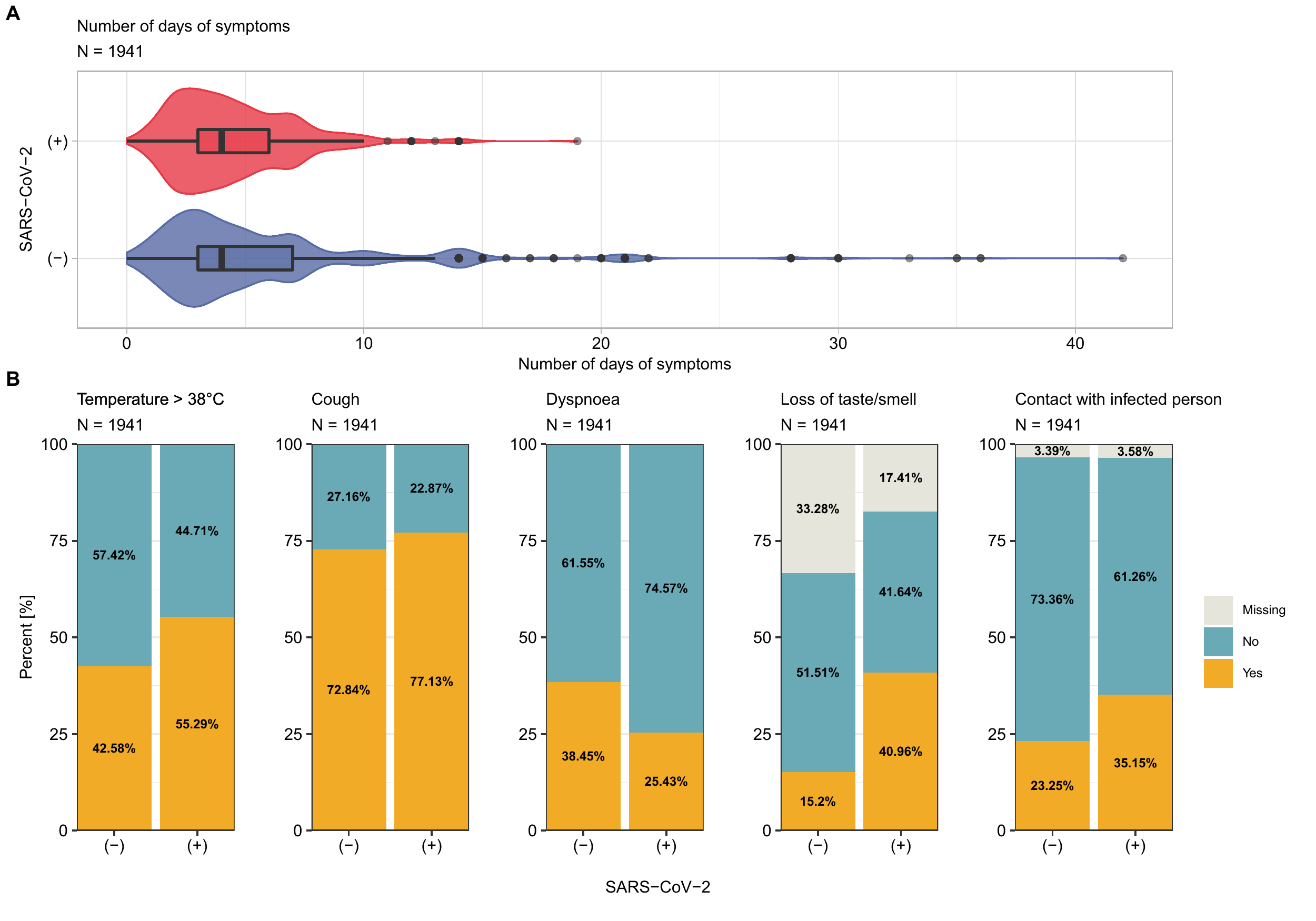}
	\caption{Distribution of exemplary attributes. Panel A shows violin plots of number of days of symptoms. Panel B shows stack bar plots for discrete attributes commonly referred to SARS-CoV-2 infection. }
	\label{fig:char}       
\end{figure}

The list of data features was reduced to 16 and it contained: sex, contact with infected person, number of days of symptoms, temperature $>38$\textsuperscript{o}C, maximal temperature, cough, dyspnoea, muscle aches, loss of smell or taste, sore throat, headache, dizziness, skin reactions, temperature (measured at the hospital), saturation, and age. 
Characteristics for exemplary attributes commonly referred as related with SARS-CoV-2 infection are presented in Fig.
\ref{fig:char}.

\section{Classification approaches}
\label{sec:classification}

Within the conducted research, two classifiers developed by two separate research groups were verified. 
It was decided that one of the model generation methods would be logistic regression \cite{mccullagh1989generalized} representing statistical approaches, while the second would be the XGBoost method \cite{10.1145/2939672.2939785} implementing Gradient Boosting model that is a leading data-driven machine learning approach. The R implementation of generalized linear models fitting for logistic regression and the \textit{xgboost} package implementing XGBoost method in the R \cite{Renv} environment were used in the analysis.

The initial data set was common to each of the approaches, however, the models were generated in separate processes.
Therefore, different data preparation and feature selection steps could be applied and hence different splits into training and test data within the model generation process were possible. However, both generated models were evaluated on a common test data set and additional collection of data available online \cite{2020covidclinicaldata}. 

As a part of the undertaken task, it was assumed that the positive examples denoted as COVID-19(+), were SARS-CoV-2 infected patients who had symptoms of COVID-19 disease. Negative examples denoted as COVID-19(--), were patients who were not infected by SARS-CoV-2 and suffer from a disease other than COVID-19.

The goal of the analysis was to derive a classification tool able to support the filtration of COVID-19(--) patients within the population of sick ones to narrow down the number of patients who need further molecular testing. Therefore, it can be said that false negative examples are more costly in this analysis than false positive ones and, using classification quality measures, that it is more important to achieve high sensitivity value than high specificity. However, when presenting the public health solutions and the results of medical analysis, the positive predictive values (PPV) and negative predictive values (NPV) 
are applied as they are commonly used to describe the performance of a screening and diagnostic test\cite{npv_ppv__reas}. As part of the research, Weighted Harmonic Mean (WHM) of NPV and PPV served in the classifier optimisation process: 
\begin{equation} \label{eq:whm}
    WHM = \left( \frac{w}{NPV} + \frac{1-w}{PPV} \right)^{-1} .
\end{equation}
In diagnostic test the weight can be equally distributed between NPV and PPV. However, here the screening test is applied, thus the importance of NPV should be higher than of PPV to maximally reduce the number of undetected COVID-19(+) patients (false negatives). This should be reflected in a value of the weight $w \in [0, 1]$.

\subsection{Logistic regression based approach}
\label{sec:class_logreg}

For the features remaining after the initial data selection, the effect size was calculated (Cram\'er's V and Rank Biserial Correlation for discrete and continuous features respectively) \cite{cramer1946mathematical,cohen2013statistical}. Only features with at least small effect were preserved for further analysis. Hence, during the logistic regression model generation five attributes were considered: i.e. contact with infected person, number of days of symptoms, temperature$>38$\textsuperscript{o}C, loss of taste/smell, and dyspnoea. Additionally, for those attributes all possible pair interactions were constructed (intersections $\cap$; logical AND) e.g. occurrence of temp.$>38$\textsuperscript{o}C and dyspnoea simultaneously will result in logical 1 for interaction variable. When at least one of the attributes is 0, the interaction value is numerically 0.

Finally, logistic regression approach does not handle missing values, so patients with incomplete information regarding five considered features were rejected. However, for the "contact with infected person" variable, the lack of an answer was equivalent to negation, so all missing values were replaced with negative responses. As a result, the training data set consisted of 572 examples, 185 of which were positive (COVID-19(+)) and 387 were negative (COVID-19(--)). 

The steps of the analysis presented in the consecutive paragraphs were repeated 100 times within the Multiple Random Cross-Validation (MRCV) process.
Patients left after filtration were randomly divided into training and validation sets in equal proportions (50:50). The division was balanced in terms of COVID-19 diagnosis as well as with regard to the considered set of features so that, for instance, patients with a particular symptom were not underrepresented in train or test set.

Next, the logistic regression model was built with the forward feature selection method. Therefore, in each step new attribute was added to the model based on the selection criterion, which in this case was the Bayes Factor. New attributes were added until the Bayes Factor value decreased below 1, which is described as a “barely worth mentioning evidence” \cite{wagenmakers2007practical},\cite{jeffreys1998theory}. The attributes were either previously selected five features or their pairs' interactions.

As the logistic regression model provides the probability that an observation belongs to a particular category (in this case COVID-19(+)), the cut-off value of the probability must be determined to classify each patient. 
The cut-off probability was identified in a way to maximize the WHM (\ref{eq:whm}) with a weight $w$ equal to 0.85. 
The enhanced importance of NPV reduces the risk of false negative observations' occurrence for the screening test purposes. Moreover, applied weight reflects standards in design medical screening test were NPV$>$90 and PPV$>$30 is expected\cite{scr_test}. For medical reasons, it is crucial to avoid COVID-19(+) patients exclusion from further diagnosis procedures, while molecular testing for SARS-CoV-2 infection of some COVID-19(-) patients is both acceptable and inevitable. The cut-off value could only be selected from the interval 0.1-0.9.
Finally, the quality of prediction for the training and validation sets were calculated.

After the MRCV procedure, the feature ranking was prepared based on the feature significance in the model and model’s quality characterized by the WHM (\ref{eq:whm}) with a weight $w$ equal to 0.85.
The forward feature selection was again applied for the final model building. However, this time the feature ranking determined the order in which attributes were added to the model. During each step, the data set was divided into training and validation part in a 50:50 proportion 100 times and quality was estimated for each division.
The model with the highest average WHM value for validation sets was selected as the final one. The cut-off probability was identified in a way to maximize the WHM value.
The parameters of the final model are presented in Table \ref{tab:logreg_params} and its performance calculated on the whole train data set is presented in Table \ref{tab:logreg_eval_train_para}.

\begin{table}[!htb]
	\caption{Parameters of the final logistic regression classifier.}
	\label{tab:logreg_params}
	\centering
	\begin{tabular}{p{100pt}cccc}
    	\hline
	&	Estimate	&	Std. Error	&	p-value & OR	\\
	    \hline
(Intercept)	&	-0.9299	&	0.2287	&	4.77E-05 & -\\
Days of symptoms	&	-0.1720	&	0.0486	&	3.99E-04 &	0.8420\\
Loss of smell or taste	&	1.4948	&	0.3729	&	6.11E-05 & 4.4584	\\
Contact with infected person	&	1.1546	&	0.2687	&	1.73E-05 &	3.1728\\
Days of symptoms AND Loss of smell or taste	&	0.0112	&	0.0593	&	8.51E-01 &	1.0113\\
Loss of smell or taste AND Contact with infected person 	&	-0.9736	&	0.4135	&	1.86E-02 &	0.3777\\
Days of symptoms AND Temp. $>38$\textsuperscript{o}C 	&	0.0763	&	0.0374	&	4.12E-02 &	1.0793\\
	    \hline
	\end{tabular}
\end{table}

\begin{table}[!htb]
	\caption{Logistic regression evaluation results - classification results for final model on the whole train set.}
	\label{tab:logreg_eval_train_para}
	\centering
	\begin{tabular}{cccc}
    	\hline
		NPV	&	PPV	&	Sensitivity	&	Specificity	\\
	    \hline
    1.0000	&	0.3551	&	1.0000	&	0.1318	\\

	    \hline
	\end{tabular}
\end{table}

\subsection{Gradient Boosting based approach}
\label{sec:class_gboost}

This approach was designed to use all available attributes and binary interactions between binary features. If two features $a$ and $b$ were binary ones then a new binary feature $a \cup b$ was created and its value was calculated by applying logical \textit{OR} operation.

The next steps of the analysis aimed in limiting the number of features involved in the classification in order to better explain the decisions made by the model.
For this purpose, all information related to the measurement of body temperature was removed from the set of features, except for the binary feature indicating whether the patient had a temperature greater than 38\textsuperscript{o}C. This step was motivated by no significant differences between the values of temperature and maximum temperature in both studied classes. Besides, the numerical variables that the XGBoost algorithm prefers to construct the tree were removed in this way.

Moreover, further feature selection based on 100 draws of a subset of the available data was performed. Each draw selected at least 60\% of the examples and 60\% of the attributes. On such a limited data set, the XGBoost algorithm was run in the 10 times five-fold cross-validation mode and the ranking of the importance of features in the trained classifier was recorded.
Then, the average position of a feature in the ranking was calculated. This average ranking consisted of only those features that occurred at least 60 times in the selected subsets. In this way, the set $A$ consisting of all features presented in Table \ref{tab:xgb_features} and their interactions (OR) between binary features was obtained. The value of \textit{sum\_of\_symptoms} feature was calculated on the basis of 11 disease symptoms included in the training set.

\begin{table}[!htb]
	\caption{Average position in the ranking of features created for XGBoost classifier.}
	\label{tab:xgb_features}
	\centering
	\begin{tabular}{lc}
    	\hline
    	Feature name	&	Average ranking position    \\
    	\hline
Days of symptoms	&	1.1	\\
No. of symptoms	&	2.5	\\
Loss of smell or taste	&	2.6	\\
Contact with infected person	&	4.6	\\
Temp. $>38$\textsuperscript{o}C	&	4.5	\\
Cough	&	5.3	\\
Muscle aches	&	5.3	\\
Dyspnoea	&	5.3	\\
	    \hline
	\end{tabular}
\end{table}

The steps of analysis presented in the consecutive paragraphs were repeated 100 times within the MRCV process.
Patients left after filtration were randomly divided into training and validation sets in 60:40 proportions. Within this procedure the optimisation process based on forward feature selection was performed.
The $WHM$ (\ref{eq:whm}) evaluation measure was used in the optimisation process, and the weight value was set to $w = 0.7$ to enhance importance of NPV and reduce in this way the risk of false negative observations’ occurrence, as the method was developed for the screening test purposes.

In order to present the complete process of building the classifier we denote by $XGB(A, t, m)$ the average value of evaluation measure $m$ calculated on the basis of the results of MRCV procedure of the XGBoost algorithm built on the basis of the set of features $A$. If $t = tr$, it means that the average value of $m$ is given on the training sets, if $t = ts$, it means that the average value of $m$ is given on the test sets. In each of the 100 experiments on the basis of which the mean value of $m$ is calculated, an internal optimization of the classification threshold to the COVID-19(+) class is carried out.

Initially, the ability to classify each individual basic feature from the $A$ set is verified. This means that for each feature $a \in A$ the scoring $XGB(\{a\}, ts, WHM)$ is calculated.
The feature with the highest scoring is selected as the best one. This feature is the first one in the set of selected features denoted as $B$.
Next, further features $a \in A\setminus B$ are added to $B$.

The process of adding features to the set of selected features $B$ is carried out as long as:
\begin{equation}
\begin{array}{l}
XGB(B, ts, WHM) < XGB(B \cup \{a\}, ts, WHM) \\
and \\
XGB(B, tr, WHM) \leq XGB(B \cup \{a\}, tr, WHM) ,
\end{array}
\end{equation}
or
\begin{equation}
\begin{array}{l}
XGB(B, ts, WHM) \leq XGB(B\cup \{a\}, ts, WHM) \\
and \\
XGB(B, tr, WHM) < XGB(B\cup \{a\}, tr, WHM) .
\end{array}
\end{equation}

After the stage of expanding the set of features, the stage of pruning redundant features can be performed. 
Feature $b$ is removed from set $B$ if its removal does not decrease (or it increases) the value of $XGB(B, ts, WHM)$, precisely:
\begin{equation}
XGB(B \setminus \{b\}, ts, WHM) \geq XGB(B, ts, WHM).
\end{equation}

The resulting set of features used further for classifier training consisted of the following seven features:
Contact with infected person OR Loss of smell or taste, Loss of smell or taste OR Temp. $>38$\textsuperscript{o}C, Days of symptoms, Cough OR Loss of smell or taste, Dyspnoea OR Contact with infected person, Temp. $>38$\textsuperscript{o}C, Cough OR Contact with infected person.

After selecting the features, the next step in the analysis was optimisation of the XGBoost algorithm parameters. It was carried out using the autoxgboost \cite{autoxgboost} R package. This time the F1 measure was used as the optimization criterion in order to increase Sensitivity and PPV value of the final classifier.

The performance of the model obtained in this way, that was calculated on the whole train data set is presented in Table \ref{tab:xgb_eval_train_para}.

\begin{table}[!htb]
	\caption{XGBoost evaluation results - classification results for final model on the whole train set.}
	\label{tab:xgb_eval_train_para}
	\centering
	\begin{tabular}{cccc}
    	\hline
    NPV	&	PPV	&	Sensitivity	&	Specificity	\\
	    \hline
    0.952	&	0.509	&	0.968	&	0.402	\\
	    \hline
	\end{tabular}
\end{table}

\section{Evaluation of created models}
\label{sec:results}

Both created classification models presented in Section \ref{sec:classification} were finally evaluated on additional data sets. The characteristic of the first test data set was presented in Section \ref{sec:data}. This data set was created as a result of a data split into training and test data (Table \ref{tab:data_char_quantity_add}) and it consists of 577 examples. This data set will be referred further as \textit{PL}.

The second data set was presented in Section \ref{sec:related_work} as the data set available online \cite{2020covidclinicaldata}. 
This data set required an initial transformation which consisted in removing all patients who did not show any of the diagnostic symptoms used by the generated models and available in this collection (contact with infected person, number of days of symptoms, temperature $>38$\textsuperscript{o}C, cough, dyspnoea, muscle aches, loss of smell or taste, sore throat, headache).
The characteristics of the obtained data set are presented in Table \ref{tab:us_char}.
This data set will be referred further as \textit{US}.

\begin{table}[!htb]
	\caption{Quantitative characteristics of the \textit{US} test data set.}
	\label{tab:us_char}
	\centering
	\begin{tabular}{lccc}
    	\hline
	    &   COVID-19(--)	&	COVID-19(+)	&	Total	\\
	    \hline
	    	&	406	&	8	&	414	\\	    
	    \hline
	\end{tabular}
\end{table}

The results of applying the XGBoost classifier to the \textit{PL} test data set are presented in Tables \ref{tab:xgb_conf_matrix_PL1_full} and \ref{tab:perf_PL1}.

\begin{table}[!htb]
	\caption{XGBoost results on the \textit{PL} test set - confusion matrix.}
	\label{tab:xgb_conf_matrix_PL1_full}
	\centering
	\begin{tabular}{llcc}
    	\hline
	&   &	\multicolumn{2}{c}{Reference}			\\
	&	&	COVID-19(--)	&	COVID-19(+)	\\
	    \hline
\multirow{2}{*}{Predicted} & COVID-19(--)	&	124	&	23	\\
    &   COVID-19(+)	&	269	&	161	\\
	    \hline
	\end{tabular}
\end{table}

The final logistic regression model, presented in Table \ref{tab:logreg_params}, was tested on the reduced \textit{PL} data set. In this set, 538 examples had full information for model features what is required by logistic regression classifier. The logistic regression model performance on this data set is presented in Tables \ref{tab:logreg_conf_matrix_PL1} and \ref{tab:perf_PL1}.

\begin{table}[!htb]
	\caption{Logistic regression results on the \textit{PL} test set without examples containing missing values - confusion matrix.}
	\label{tab:logreg_conf_matrix_PL1}
	\centering
	\begin{tabular}{llcc}
    	\hline
	&   &	\multicolumn{2}{c}{Reference}			\\
	&	&	COVID-19(--)	&	COVID-19(+)	\\
	    \hline
\multirow{2}{*}{Predicted} & COVID-19(--)	&	41	&	4	\\
    &   COVID-19(+)	&	322	&	171	\\
	    \hline
	\end{tabular}
\end{table}

Additionally, the XGBoost classifier was applied to the reduced \textit{PL} test data set, in which there are no missing values. In this way it was possible to compare the quality of both models.
These results of XGBoost are presented in Tables \ref{tab:xgb_conf_matrix_PL1} and \ref{tab:perf_PL1}. 

\begin{table}[!htb]
	\caption{XGBoost results on the \textit{PL} test set without examples containing missing values - confusion matrix.}
	\label{tab:xgb_conf_matrix_PL1}
	\centering
	\begin{tabular}{llcc}
    	\hline
	&   &	\multicolumn{2}{c}{Reference}			\\
	&	&	COVID-19(--)	&	COVID-19(+)	\\
	    \hline
\multirow{2}{*}{Predicted} & COVID-19(--)	&	117	&	23	\\
    &   COVID-19(+)	&	244	&	152	\\
	    \hline
	\end{tabular}
\end{table}

\begin{table}[!htb]
	\caption{Classification quality of the logistic regression and XGBoost models on the \textit{PL} test data set.}
	\label{tab:perf_PL1}
	\centering
	\begin{tabular}{lccc}
    	\hline
     &   XGBoost &   Logistic regression    &    XGBoost   	\\
    &   Full data set   &   Reduced data set    &   Reduced data set    \\
    	\hline
Sensitivity	&	0.875	&	0.977	&	0.869	\\
Specificity	&	0.316	&	0.113	&	0.324	\\
PPV	&	0.374	&	0.347	&	0.384	\\
NPV	&	0.844	&	0.911	&	0.836	\\
	    \hline
	\end{tabular}
\end{table}

Second evaluation of both generated classifiers was performed on \textit{US} data set. The XGBoost model performance on this data set is presented in Tables \ref{tab:xgb_conf_matrix_US_full} and \ref{tab:perf_US}. Again, the logistic regression model was tested on the reduced \textit{US} data set, from which examples with missing values were removed, and it consisted of 337 examples. The logistic regression performance on this data set is presented in Tables \ref{tab:logreg_conf_matrix_US} and \ref{tab:perf_US}. 
The results of XGBoost tested on the reduced \textit{US} data set are presented in Tables \ref{tab:xgb_conf_matrix_US} and \ref{tab:perf_US}. 

\begin{table}[!htb]
	\caption{XGBoost results on the \textit{US} test data set - confusion matrix.}
	\label{tab:xgb_conf_matrix_US_full}
	\centering
	\begin{tabular}{llcc}
    	\hline
	&   &	\multicolumn{2}{c}{Reference}			\\
	&	&	COVID-19(--)	&	COVID-19(+)	\\
	    \hline
\multirow{2}{*}{Predicted} & COVID-19(--)	&	211	&	3	\\
    &   COVID-19(+)	&	195	&	5	\\
	    \hline
	\end{tabular}
\end{table}

\begin{table}[!htb]
	\caption{Logistic regression results on the \textit{US} test data set without examples containing missing values - confusion matrix.}
	\label{tab:logreg_conf_matrix_US}
	\centering
	\begin{tabular}{llcc}
    	\hline
	&   &	\multicolumn{2}{c}{Reference}			\\
	&	&	COVID-19(--)	&	COVID-19(+)	\\
	    \hline
\multirow{2}{*}{Predicted} & COVID-19(--)	&	65	&	1	\\
    &   COVID-19(+)	&	213	&	5	\\
	    \hline
	\end{tabular}
\end{table}

\begin{table}[!htb]
	\caption{XGBoost results on the \textit{US} test data set without examples containing missing values - confusion matrix.}
	\label{tab:xgb_conf_matrix_US}
	\centering
	\begin{tabular}{llcc}
    	\hline
	&   &	\multicolumn{2}{c}{Reference}			\\
	&	&	COVID-19(--)	&	COVID-19(+)	\\
	    \hline
\multirow{2}{*}{Predicted} & COVID-19(--)	&	145	&	2	\\
    &   COVID-19(+)	&	133	&	4	\\
	    
	    \hline
	\end{tabular}
\end{table}

\begin{table}[!htb]
	\caption{Classification quality of the logistic regression and XGBoost models on the \textit{US} test data set.}
	\label{tab:perf_US}
	\centering
	\begin{tabular}{lccc}
    	\hline
    &   XGBoost &   Logistic regression    &    XGBoost   	\\
    &   Full data set   &   Reduced data set    &   Reduced data set    \\

    	\hline

Sensitivity	&	0.625	&	0.833	&	0.667	\\
Specificity	&	0.520	&	0.234	&	0.522	\\
PPV	&	0.025	&	0.023	&	0.029	\\
NPV	&	0.986	&	0.985	&	0.986	\\

	    \hline
	\end{tabular}
\end{table}

\section{Discussion}
\label{sec:discussion}

The two presented classifiers developed in independent groups use a similar set of features. The logistic regression model uses the features listed in section \ref{sec:class_logreg} in Table \ref{tab:logreg_params}. The XGBoost model uses the features listed in section \ref{sec:class_gboost}.
The XGBoost model uses two more base features compared to logistic regression, i.e. dyspnoea and cough.

It is worth noting that both models are largely based on compound features that were created in various ways. In the logistic regression model, interactions are modeled by the logical operator AND, and in the XGBoost model by OR.
This is partly due to conducting research by two independent teams, but it also reflects well how different methods can recreate interactions. Logistic regression does not directly model the conjunction of the feature values, and the alternative of feature values cannot be easily modeled in decision trees.

The obtained classification results show that the NPV of both methods is acceptable for screening test, in particular the NPV of the logistic regression model is above 0.90 on the \textit{PL} and \textit{US} sets. Both models have similar PPV values. The logistic regression model is characterized by higher values of NPV and Sensitivity, and XGBoost is characterized by higher values of Specificity.
This is partly due to the model optimization method. Besides, it confirms the thesis that depending on the user's preferences expressed in this article by the weight value in formula (\ref{eq:whm}), it is possible to obtain classifiers with different levels of PPV and NPV.

The relation between parameter values and the decision represented by XGBoost models is difficult to interpret by a human. Logistic regression gives the possibility to estimate the adjusted odds ratio (Table \ref{tab:logreg_params}), which gives direct and intuitive interpretation of obtained classifier parameters. However, to give a unified comparison of both approaches the rpart decision trees \cite{Therneau2109rpart} were created to provide explanation between symptoms and the outcome predicted by the models. 

Having two sets of basic features selected for each model, decision trees were generated to explain the behaviour of their representative models. First, two training sets were prepared. The data sets consisted of patients with symptoms (Sick) representing the attributes which were used by the models and each patient was assigned the decision given by the model. Then, based on these two separate training sets two rpart decision trees were generated. Finally, it was evaluated how well the decision trees approximate the decisions made by the considered models. For each patient the decision given by the model was compared with the decision given by its rpart approximation.
For the decision tree generated for the logistic regression model approximation the resulting Balanced Accuracy (BAcc) value was 0.967 and for the decision tree generated for the XGBoost model approximation the resulting BAcc value was 0.937. Based on such very high BAcc values it was possible to use the trees to explain the relation between symptoms and models outcome.  

Besides, similar decision trees using training sets based on selected feature interactions were generated. 
However, since the representation of a tree based on simple (basic) features can be better understood it was decided to include a simpler version in the manuscript. 

\begin{figure}[!htb]
	\includegraphics[width=1\textwidth]{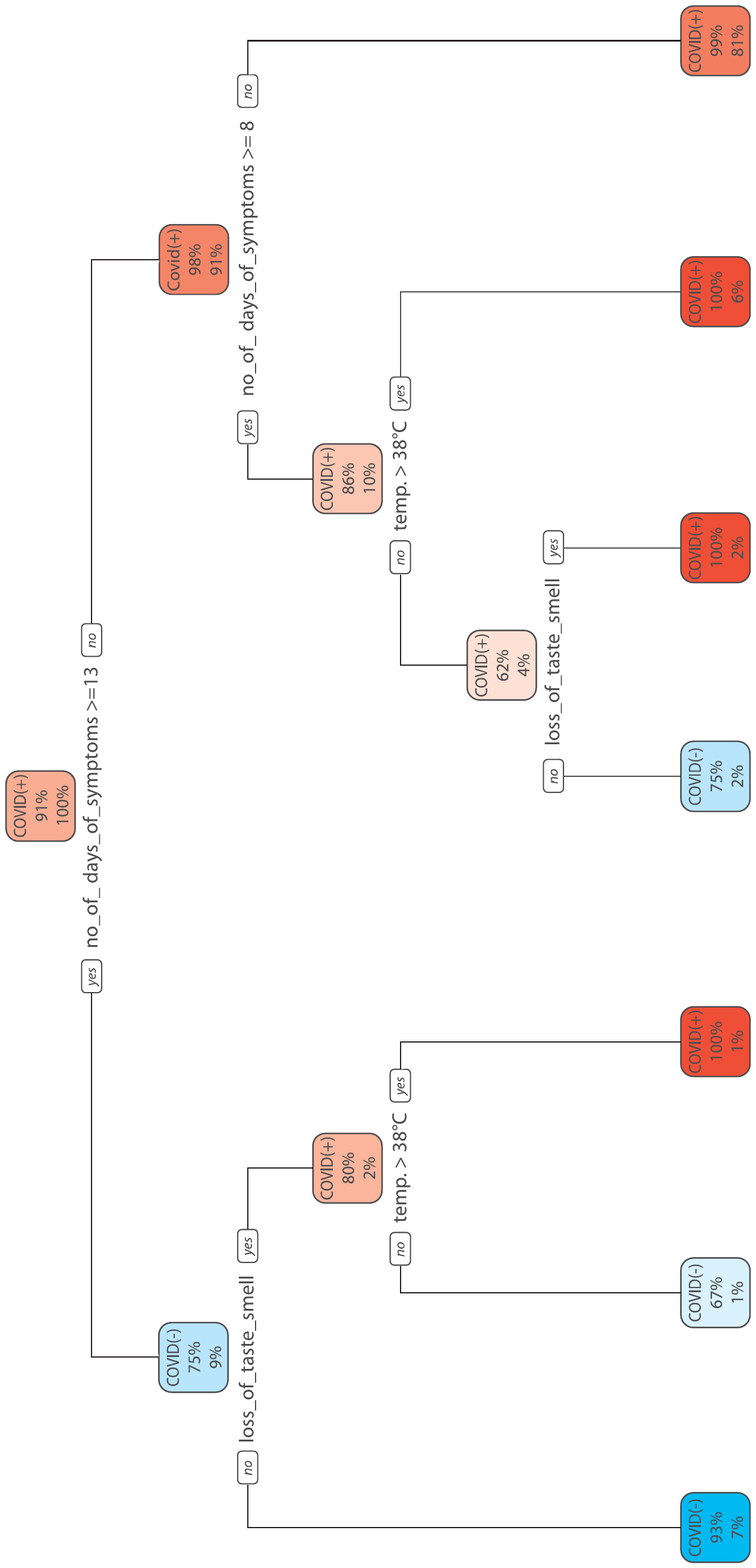}
	\caption{Decision tree generated to explain the behaviour of logistic regression classifier.}
	\label{fig:dec_tree_logreg}       
\end{figure}

\begin{figure}[!htb]
	\includegraphics[width=1\textwidth]{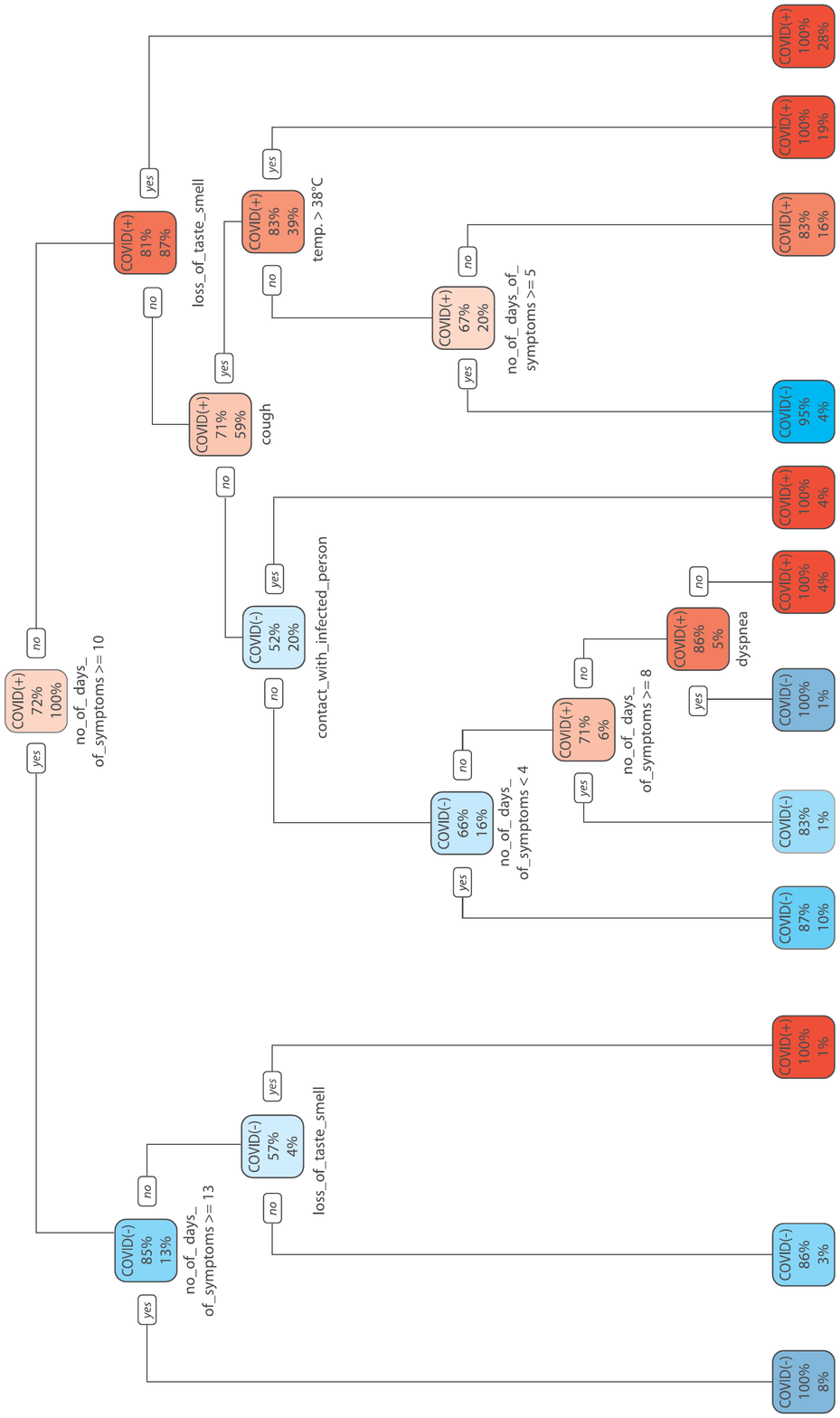}
	\caption{Decision tree generated to explain the behaviour of XGBoost classifier.}
	\label{fig:dec_tree_xgb}       
\end{figure}

Generated decisions trees are presented in the Figures \ref{fig:dec_tree_logreg} and \ref{fig:dec_tree_xgb}. 
Colours on the Figures represent classes - COVID-19(+) is represented by red, COVID-19(--) by blue. The more saturated the color of a node is, the greater the dominance of examples from a given class in this node.
The first row of rpart tree node provides the information about the majority class of examples covered by that node, the second provides the percentage of the cases from the node that are classified by the approximated model into the majority class, and the third row provides the percentage of all cases covered by that node. Branches of the tree represent symptoms and their presence or lack thereof (\textit{yes} means that patient has a symptom and \textit{no} that a symptom was not observed). Please note that since the trees were built to approximate the models, their leaves do not represent the patient outcome but the decision provided by the model based on the particular set of symptoms. 
Decision trees can be used to explain which combinations of symptoms and their values are taken into account while predicting diagnosis. 
For example, it can be noted that in case of both models, the long time of having symptoms \textit{no\_of\_days\_of\_symptoms $>$= 13} will results in providing negative decision (COVID-19(--)), unless the patient observes the  \textit{loss\_of\_taste\_smell}.  

It is known from the literature \cite{Meng_COVID_2020,Pierron_Smell_2020}, that loss of taste or smell is frequently used as an early indicator of SARS-Cov-2 infection and therefore it is not surprising that having it results in the positive decision (COVID-19(+)) in almost all cases for both classifiers. The only exception of this is a case of the tree generated for the logistic regression model, where having more than 13 days of symptoms together with loss of taste or smell and temperature below 38\textsuperscript{o}C results in the negative decision. However, in this case, this concerns only 67\% of cases from the leaf covering only 1\% of examples. Another symptom very strongly related to the positive outcome is having a high temperature (\textit{temp.$>$38\textsuperscript{o}C}). Similarly, like in the previous cases, having significantly elevated temperature predominantly results in the positive decision of the classifiers.

In general, the decision tree approximating the logistic regression model is less complex and allows easy interpretation of the model's decision making. The decision making model can be expressed in 7 rules containing from 2 to 4 conditions (nodes). The tree approximating the XGBoost model is more complex.
In this case the decision model consists of 12 rules with 2 to 7 conditions. It is interesting to analyse the description of the rules covering the largest number of examples (i.e. rules built on the basis of tree paths from the root to the extreme right and left leaves).

In case of logistic regression, a large part of the COVID-19(+) decisions is made based on the number of days the patient has symptoms. If this number is less than 8, the patient is classified as COVID-19(+). However, if this number is greater than 13 and there is no loss of smell or taste, the patient is classified as COVID-19(--). The first rule seems debatable, however, it can be explained in such a way that in the era of the pandemic developing - in Poland a bit later than in Western Europe - people who observed the basic symptoms associated in the media with the possibility of COVID-19(+), decided to test themselves for SARS-CoV-2.
It should also be remembered that the described tree represents the way of decisions made by the logistic regression model and not the actual data set, While this model was optimized according to WHM, where $w = 0.85$, which means that high NPV values were preferred. Thus, the rule is intuitive.

The extreme rules of the tree approximating the XGBoost model are more intuitive but also cover fewer examples. The main rule classifying patients as COVID-19(+) depends on the number of days of symptoms ($<10$) and loss of taste or smell (yes).
The main rule to classify patients as COVID-19(--) is based only on the number of days of persistent symptoms and if it is long enough ($>=13$), the patient is classified as COVID-19(--).
The remaining rules describe smaller sets of examples, and the decision is made on the basis of more detailed information concerning e.g. contact with an infected person, the presence of temperature above 38\textsuperscript{o}C, cough and dyspnoea (the last two features are used only in the XGBoost model).

\section{DECODE service}
\label{sec:app}

DECODE is a symptom checker tool to assist patients and family doctors in preliminary screening and early detection of COVID-19. 

In DECODE service a patient has the opportunity to fill out a questionnaire regarding his/her health condition and obtain a preliminary assessment of the possibility of being sick with COVID-19 and, consequently, the need to be tested for SARS-CoV-2.

\begin{figure}[!htb]
	\includegraphics[width=0.9\textwidth]{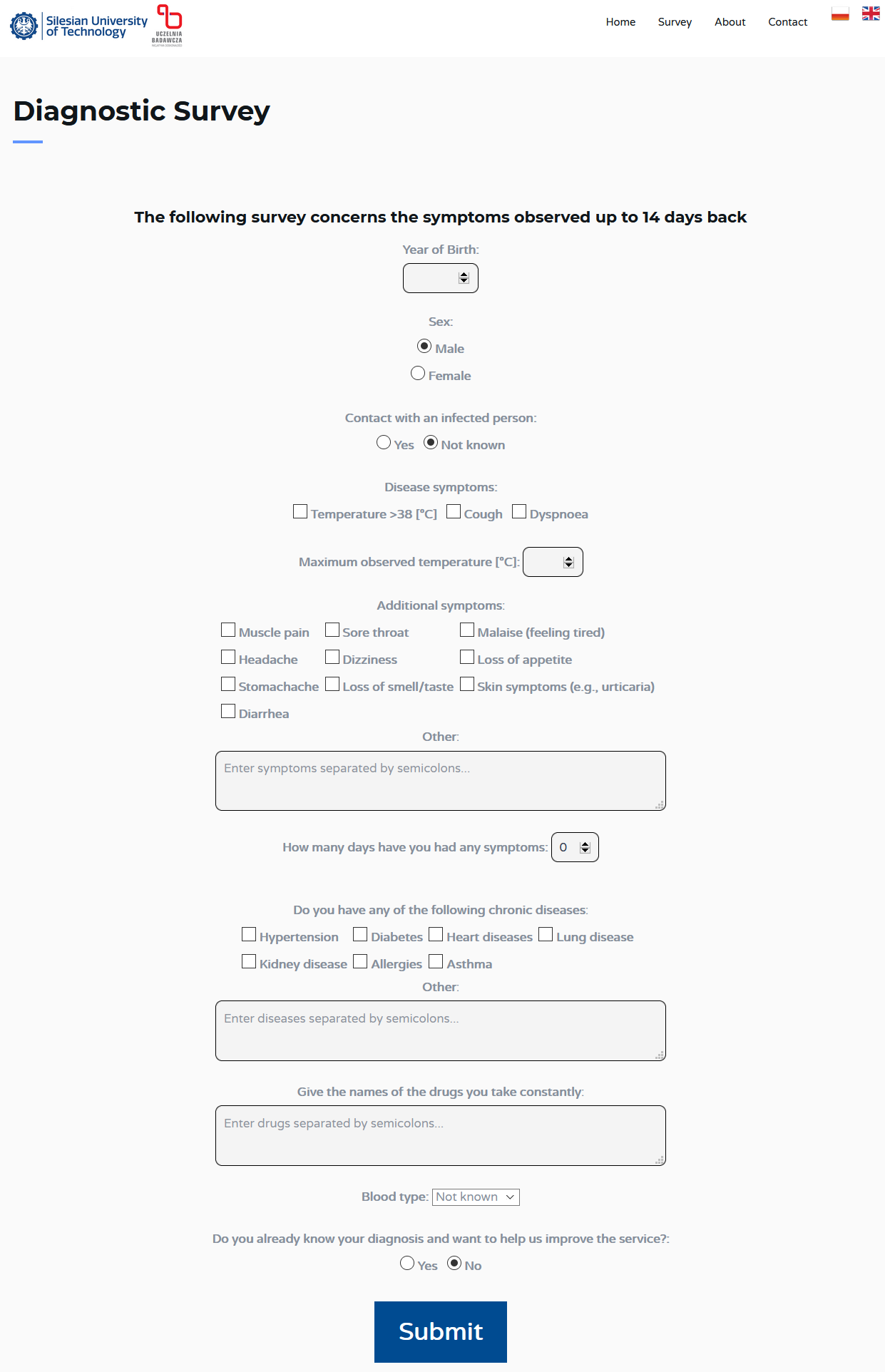}
	\caption{The example of questionnaire}
	\label{fig:questionnaire}       
\end{figure}

The questions (Fig. \ref{fig:questionnaire}) cover many problem areas related to possible COVID-19 disease. In addition to the year of birth, sex, occurrence of signs that are typical for coronavirus, such as fever over 38 degrees of Celsius, cough, dyspnoea, loss of smell or taste, etc., the patient may add other disturbing symptoms and state how long they last. Besides, the person completing the questionnaire has the option of providing the names of chronic diseases – if any, and the names of the medications that are constantly taken. Blood type is also taken into account. In order to avoid possible inconsistencies, the questionnaire fields are validated. For example, if a patient checks any symptoms, the length of their presence must be greater than 0.

After completing and sending the questionnaire, the patient receives suggestions in the form of Negative/Positive and information which of the reported symptoms and to what extent influenced the result. Moreover, the system informs users that obtained results are only prediction and only medical test can fully confirm infection of SARS-CoV-2.

The link that allows the patient to return to the survey is active for 14 days. Using this link the patient can, for example, report the next symptoms, if they occur. In particular, when the patient knows what the PCR test results are, he/she can share this information and help improve the service.

DECODE is available in two language versions (Polish and English) and is integrated with the CIRCA diagnostic service which was also developed at the Silesian University of Technology to support and accelerate COVID-19 imaging diagnostics.

\section{Conclusions}
\label{sec:conclusions}

The paper presents the results of the study on classifiers tuned towards achieving an assumed acceptable threshold of negative predictive values during classification (screening) of patients suffering from COVID-19 disease.

Based on the feature set describing the disease symptoms, it was possible to obtain classifiers that met the expectations concerning the detection of patients with COVID-19 disease. This has been achieved at the expense of a relatively low Positive Predictive Value. The classifiers' efficiency was verified on the data set obtained as part of the presented research, in the last period of collecting the questionnaires and on the data set available online. 

Thanks to an approximation of obtained classification models by a decision tree, it was possible to explain – using the set of basic features – what features and their values are the basis of decisions taken by the classifiers.

The obtained classification models provided the basis for the development of the DECODE service (decode.polsl.pl) allowing the doctors for preselection for further diagnostics in the event of the overwhelming increase in disease.

An inseparable part of the paper is a data set describing over 3100 patients with verified (positively or negatively) SARS-CoV-2 infection. The set includes: asymptomatic patients with SARS-CoV-2 infection, patients with COVID-19 disease (referred to in the paper as COVID-19(+)), as well as symptomatic patients but without SARS-CoV-2 infection (suffering from other diseases, e.g. influenza, denoted as COVID-19(--)).
In addition to information about symptoms, the data set includes information on comorbidities (e.g. hypertension, diabetes, etc.). However, comorbidities were not analysed in the studies described in this paper. The data set is available to a wide group of researchers and it is a significant data repository describing COVID-19 symptoms in the Slavic population.

Further work will focus on improving classifiers to increase their specificity (PPV and Specificity). This goal will be achieved by gathering a larger set of examples obtained from the DECODE webservice users and through cooperation with a greater number of hospitals.
The developed classification models will be periodically tuned and verified on newly emerging data.
Another intended research will start, in cooperation with hospitals, an analysis of the severity of the course of the COVID-19 disease, as well as a survival analysis of patients with a specific set of comorbidities. 
Besides, it is intended to study the course of the COVID-19 disease depending on the medications taken by patients (the DECODE webservice enables gathering this information).

\section*{Acknowledgements}
\label{sec:acknowledgements}
This work is supported by the Silesian University of Technology grant for Support and Development of Research Potential[JH, JT, MKo, MB, PF, AG, MKa, JM, AP, AW, JZ, JP, MS]. We would like to thank Bożena W{\l}ostowska, Sylwia Wróbel, Iwona Kostorz, Katarzyna Fabiś for support in process of rewriting patient surveys.


\bibliography{mybibfile}

\begin{thebibliography}{10}
\expandafter\ifx\csname url\endcsname\relax
  \def\url#1{\texttt{#1}}\fi
\expandafter\ifx\csname urlprefix\endcsname\relax\def\urlprefix{URL }\fi
\expandafter\ifx\csname href\endcsname\relax
  \def\href#1#2{#2} \def\path#1{#1}\fi

\bibitem{caballe2020machine}
N.~C. Caball{\'e}, J.~L. Castillo-Sequera, J.~A. G{\'o}mez-Pulido, J.~M.
  G{\'o}mez-Pulido, M.~L. Polo-Luque, Machine learning applied to diagnosis of
  human diseases: A systematic review, Applied Sciences 10~(15) (2020) 5135.

\bibitem{COLUBRI201954}
A.~Colubri, M.-A. Hartley, M.~Siakor, V.~Wolfman, A.~Felix, T.~Sesay, J.~G.
  Shaffer, R.~F. Garry, D.~S. Grant, A.~C. Levine, P.~C. Sabeti,
  \href{http://www.sciencedirect.com/science/article/pii/S2589537019300963}{Machine-learning
  prognostic models from the 2014–16 {Ebola} outbreak: Data-harmonization
  challenges, validation strategies, and {mHealth} applications},
  EClinicalMedicine 11 (2019) 54 -- 64.
\newblock \href
  {http://dx.doi.org/https://doi.org/10.1016/j.eclinm.2019.06.003}
  {\path{doi:https://doi.org/10.1016/j.eclinm.2019.06.003}}.
\newline\urlprefix\url{http://www.sciencedirect.com/science/article/pii/S2589537019300963}

\bibitem{CHOCKANATHAN201924}
U.~Chockanathan, A.~M. DSouza, A.~Z. Abidin, G.~Schifitto, A.~Wismüller,
  \href{http://www.sciencedirect.com/science/article/pii/S001048251930006X}{Automated
  diagnosis of hiv-associated neurocognitive disorders using large-scale
  {Granger} causality analysis of resting-state functional mri}, Computers in
  Biology and Medicine 106 (2019) 24 -- 30.
\newblock \href
  {http://dx.doi.org/https://doi.org/10.1016/j.compbiomed.2019.01.006}
  {\path{doi:https://doi.org/10.1016/j.compbiomed.2019.01.006}}.
\newline\urlprefix\url{http://www.sciencedirect.com/science/article/pii/S001048251930006X}

\bibitem{GARATEESCAMILA2020100330}
A.~K. Gárate-Escamila, A.~{Hajjam El Hassani}, E.~Andrès,
  \href{http://www.sciencedirect.com/science/article/pii/S2352914820300125}{Classification
  models for heart disease prediction using feature selection and {PCA}},
  Informatics in Medicine Unlocked 19 (2020) 100330.
\newblock \href {http://dx.doi.org/https://doi.org/10.1016/j.imu.2020.100330}
  {\path{doi:https://doi.org/10.1016/j.imu.2020.100330}}.
\newline\urlprefix\url{http://www.sciencedirect.com/science/article/pii/S2352914820300125}

\bibitem{SAXENA2020182}
S.~Saxena, M.~Gyanchandani,
  \href{http://www.sciencedirect.com/science/article/pii/S193986541930551X}{Machine
  learning methods for computer-aided breast cancer diagnosis using
  histopathology: A narrative review}, Journal of Medical Imaging and Radiation
  Sciences 51~(1) (2020) 182 -- 193.
\newblock \href {http://dx.doi.org/https://doi.org/10.1016/j.jmir.2019.11.001}
  {\path{doi:https://doi.org/10.1016/j.jmir.2019.11.001}}.
\newline\urlprefix\url{http://www.sciencedirect.com/science/article/pii/S193986541930551X}

\bibitem{KAVAKIOTIS2017104}
I.~Kavakiotis, O.~Tsave, A.~Salifoglou, N.~Maglaveras, I.~Vlahavas,
  I.~Chouvarda,
  \href{http://www.sciencedirect.com/science/article/pii/S2001037016300733}{Machine
  learning and data mining methods in diabetes research}, Computational and
  Structural Biotechnology Journal 15 (2017) 104 -- 116.
\newblock \href {http://dx.doi.org/https://doi.org/10.1016/j.csbj.2016.12.005}
  {\path{doi:https://doi.org/10.1016/j.csbj.2016.12.005}}.
\newline\urlprefix\url{http://www.sciencedirect.com/science/article/pii/S2001037016300733}

\bibitem{Wynantsm1328}
L.~Wynants, B.~Van~Calster, G.~S. Collins, R.~D. Riley, G.~Heinze, E.~Schuit,
  M.~M.~J. Bonten, D.~L. Dahly, J.~A.~A. Damen, T.~P.~A. Debray, V.~M.~T.
  de~Jong, M.~De~Vos, P.~Dhiman, M.~C. Haller, M.~O. Harhay, L.~Henckaerts,
  P.~Heus, N.~Kreuzberger, A.~Lohmann, K.~Luijken, J.~Ma, G.~P. Martin, C.~L.
  Andaur~Navarro, J.~B. Reitsma, J.~C. Sergeant, C.~Shi, N.~Skoetz, L.~J.~M.
  Smits, K.~I.~E. Snell, M.~Sperrin, R.~Spijker, E.~W. Steyerberg, T.~Takada,
  I.~Tzoulaki, S.~M.~J. van Kuijk, F.~S. van Royen, J.~Y. Verbakel,
  C.~Wallisch, J.~Wilkinson, R.~Wolff, L.~Hooft, K.~G.~M. Moons, M.~van Smeden,
  \href{https://www.bmj.com/content/369/bmj.m1328}{Prediction models for
  diagnosis and prognosis of {COVID-19}: systematic review and critical
  appraisal}, BMJ 369.
\newblock \href
  {http://arxiv.org/abs/https://www.bmj.com/content/369/bmj.m1328.full.pdf}
  {\path{arXiv:https://www.bmj.com/content/369/bmj.m1328.full.pdf}}, \href
  {http://dx.doi.org/10.1136/bmj.m1328} {\path{doi:10.1136/bmj.m1328}}.
\newline\urlprefix\url{https://www.bmj.com/content/369/bmj.m1328}

\bibitem{CHRISTODOULOU201912}
E.~Christodoulou, J.~Ma, G.~S. Collins, E.~W. Steyerberg, J.~Y. Verbakel,
  B.~{Van Calster},
  \href{http://www.sciencedirect.com/science/article/pii/S0895435618310813}{A
  systematic review shows no performance benefit of machine learning over
  logistic regression for clinical prediction models}, Journal of Clinical
  Epidemiology 110 (2019) 12 -- 22.
\newblock \href
  {http://dx.doi.org/https://doi.org/10.1016/j.jclinepi.2019.02.004}
  {\path{doi:https://doi.org/10.1016/j.jclinepi.2019.02.004}}.
\newline\urlprefix\url{http://www.sciencedirect.com/science/article/pii/S0895435618310813}

\bibitem{AHAMAD2020113661}
M.~M. Ahamad, S.~Aktar, M.~Rashed-Al-Mahfuz, S.~Uddin, P.~Liò, H.~Xu, M.~A.
  Summers, J.~M. Quinn, M.~A. Moni,
  \href{http://www.sciencedirect.com/science/article/pii/S0957417420304851}{A
  machine learning model to identify early stage symptoms of {SARS-Cov-2}
  infected patients}, Expert Systems with Applications 160 (2020) 113661.
\newblock \href {http://dx.doi.org/https://doi.org/10.1016/j.eswa.2020.113661}
  {\path{doi:https://doi.org/10.1016/j.eswa.2020.113661}}.
\newline\urlprefix\url{http://www.sciencedirect.com/science/article/pii/S0957417420304851}

\bibitem{ma_ng_xu_xu_qiu_liu_lyu_you_zhao_wang}
X.~Ma, M.~Ng, S.~Xu, Z.~Xu, H.~Qiu, Y.~Liu, J.~Lyu, J.~You, P.~Zhao, S.~Wang,
  et~al., Development and validation of prognosis model of mortality risk in
  patients with {COVID-19}, Epidemiology and Infection 148 (2020) e168.
\newblock \href {http://dx.doi.org/10.1017/S0950268820001727}
  {\path{doi:10.1017/S0950268820001727}}.

\bibitem{Escobar17720}
L.~E. Escobar, A.~Molina-Cruz, C.~Barillas-Mury,
  \href{https://www.pnas.org/content/117/30/17720}{{BCG} vaccine protection
  from severe coronavirus disease 2019 ({COVID-19})}, Proceedings of the
  National Academy of Sciences 117~(30) (2020) 17720--17726.
\newblock \href
  {http://arxiv.org/abs/https://www.pnas.org/content/117/30/17720.full.pdf}
  {\path{arXiv:https://www.pnas.org/content/117/30/17720.full.pdf}}, \href
  {http://dx.doi.org/10.1073/pnas.2008410117}
  {\path{doi:10.1073/pnas.2008410117}}.
\newline\urlprefix\url{https://www.pnas.org/content/117/30/17720}

\bibitem{SWAPNAREKHA2020109947}
H.~Swapnarekha, H.~S. Behera, J.~Nayak, B.~Naik,
  \href{http://www.sciencedirect.com/science/article/pii/S0960077920303465}{Role
  of intelligent computing in covid-19 prognosis: A state-of-the-art review},
  Chaos, Solitons \& Fractals 138 (2020) 109947.
\newblock \href {http://dx.doi.org/https://doi.org/10.1016/j.chaos.2020.109947}
  {\path{doi:https://doi.org/10.1016/j.chaos.2020.109947}}.
\newline\urlprefix\url{http://www.sciencedirect.com/science/article/pii/S0960077920303465}

\bibitem{LALMUANAWMA2020110059}
S.~Lalmuanawma, J.~Hussain, L.~Chhakchhuak,
  \href{http://www.sciencedirect.com/science/article/pii/S0960077920304562}{Applications
  of machine learning and artificial intelligence for {Covid-19 (SARS-CoV-2)}
  pandemic: A review}, Chaos, Solitons \& Fractals 139 (2020) 110059.
\newblock \href {http://dx.doi.org/https://doi.org/10.1016/j.chaos.2020.110059}
  {\path{doi:https://doi.org/10.1016/j.chaos.2020.110059}}.
\newline\urlprefix\url{http://www.sciencedirect.com/science/article/pii/S0960077920304562}

\bibitem{Menni2020.04.05.20048421}
C.~Menni, A.~Valdes, M.~B. Freydin, S.~Ganesh, J.~El-Sayed~Moustafa,
  A.~Visconti, P.~Hysi, R.~C.~E. Bowyer, M.~Mangino, M.~Falchi, J.~Wolf,
  C.~Steves, T.~Spector,
  \href{https://www.medrxiv.org/content/early/2020/04/07/2020.04.05.20048421}{Loss
  of smell and taste in combination with other symptoms is a strong predictor
  of {COVID-19} infection}, medRxiv\href
  {http://arxiv.org/abs/https://www.medrxiv.org/content/early/2020/04/07/2020.04.05.20048421.full.pdf}
  {\path{arXiv:https://www.medrxiv.org/content/early/2020/04/07/2020.04.05.20048421.full.pdf}},
  \href {http://dx.doi.org/10.1101/2020.04.05.20048421}
  {\path{doi:10.1101/2020.04.05.20048421}}.
\newline\urlprefix\url{https://www.medrxiv.org/content/early/2020/04/07/2020.04.05.20048421}

\bibitem{Diaz-Quijano2020.04.05.20047944}
F.~A. Diaz-Quijano, J.~M. N.~d. Silva, F.~Ganem, S.~Oliveira, A.~L.
  Vesga-Varela, J.~Croda,
  \href{https://www.medrxiv.org/content/early/2020/08/03/2020.04.05.20047944}{A
  model to predict {SARS-CoV-2} infection based on the first three-month
  surveillance data in {Brazil}.}, medRxiv\href
  {http://arxiv.org/abs/https://www.medrxiv.org/content/early/2020/08/03/2020.04.05.20047944.full.pdf}
  {\path{arXiv:https://www.medrxiv.org/content/early/2020/08/03/2020.04.05.20047944.full.pdf}},
  \href {http://dx.doi.org/10.1101/2020.04.05.20047944}
  {\path{doi:10.1101/2020.04.05.20047944}}.
\newline\urlprefix\url{https://www.medrxiv.org/content/early/2020/08/03/2020.04.05.20047944}

\bibitem{shuja2020covid}
J.~Shuja, E.~Alanazi, W.~Alasmary, A.~Alashaikh, {COVID-19} open source data
  sets: a comprehensive survey, Applied Intelligence (2020) 1--30.

\bibitem{2020covidclinicaldata}
{Carbon Health and Braid Health}, {Coronavirus Disease 2019 ({COVID-19})
  Clinical Data Repository}, Accessed from
  \url{https://covidclinicaldata.org/.} (2020).

\bibitem{kaggle_covid}
{Einstein Data4u}, {COVID-19 clinical data Hospital collected at the Israelita
  Albert Einstein, at Sao Paulo, Brazil}, Accessed from
  https://www.kaggle.com/einsteindata4u/covid19 (2020).

\bibitem{schwab2020predcovid}
P.~Schwab, A.~D. Sch{\"u}tte, B.~Dietz, S.~Bauer, {predCOVID-19}: A systematic
  study of clinical predictive models for coronavirus disease 2019, arXiv
  preprint arXiv:2005.08302.

\bibitem{Batista2020.04.04.20052092}
A.~F. d.~M. Batista, J.~L. Miraglia, T.~H.~R. Donato, A.~D.~P.
  Chiavegatto~Filho,
  \href{https://www.medrxiv.org/content/early/2020/04/14/2020.04.04.20052092}{{COVID-19}
  diagnosis prediction in emergency care patients: a machine learning
  approach}, medRxiv\href
  {http://arxiv.org/abs/https://www.medrxiv.org/content/early/2020/04/14/2020.04.04.20052092.full.pdf}
  {\path{arXiv:https://www.medrxiv.org/content/early/2020/04/14/2020.04.04.20052092.full.pdf}},
  \href {http://dx.doi.org/10.1101/2020.04.04.20052092}
  {\path{doi:10.1101/2020.04.04.20052092}}.
\newline\urlprefix\url{https://www.medrxiv.org/content/early/2020/04/14/2020.04.04.20052092}

\bibitem{European51:online}
{European mhealth hub | Home mHealth Hub}, \url{https://mhealth-hub.org/},
  (Accessed on 10/21/2020).

\bibitem{European43:online}
European mhealth hub | {COVID-19} apps hub repository,
  \url{https://mhealth-hub.org/mhealth-solutions-against-covid-19}, (Accessed
  on 10/21/2020).

\bibitem{TheNHSCO40:online}
{The NHS COVID-19 app support website - NHS.UK},
  \url{https://www.covid19.nhs.uk/}, (Accessed on 10/21/2020).

\bibitem{ProteGO_Safe:online}
{STOP COVID – ProteGO Safe}, \url{https://www.gov.pl/web/protegosafe},
  (Accessed on 10/27/2020).

\bibitem{Radar_COVID:online}
{Radar COVID},
  \url{https://www.lamoncloa.gob.es/lang/en/gobierno/news/Paginas/2020/20200803radarcovid.aspx},
  (Accessed on 10/27/2020).

\bibitem{LocalCOV28:online}
GOV.UK, {Local COVID alert levels: what you need to know},
  \url{https://www.gov.uk/guidance/local-covid-alert-levels-what-you-need-to-know},
  (Accessed on 10/21/2020).

\bibitem{CERCACOVID:online}
Allertalom – cercacovid,
  \url{https://www.openinnovation.regione.lombardia.it/b/572/regioneaicittadiniunapppermonitorareladiffusionedelcovid},
  (Accessed on 10/27/2020).

\bibitem{WHO_ACADEMY:online}
Who academy, \url{https://www.who.int/about/who-academy}, (Accessed on
  10/27/2020).

\bibitem{healthdi61:online}
Healthdirect, healthdirect symptom checker,
  \url{https://www.healthdirect.gov.au/symptom-checker/tool/basic-details},
  (Accessed on 10/21/2020).

\bibitem{Testingf99:online}
CDC, {Testing for COVID-19},
  \url{https://www.cdc.gov/coronavirus/2019-ncov/symptoms-testing/testing.html},
  (Accessed on 10/21/2020).

\bibitem{Koronawi16:online}
A.~Inc., Apple covid-19, \url{https://covid19.apple.com/screening}, (Accessed
  on 10/21/2020).

\bibitem{COVID19R49:online}
{COVID-19 Risk Assessment}, \url{https://covid.preflet.com/en}, (Accessed on
  10/21/2020).

\bibitem{Mediktor72:online}
Mediktor, {Mediktor - AI-based medical assistant},
  \url{https://www.mediktor.com/en}, (Accessed on 10/21/2020).

\bibitem{TheHuman42:online}
The human diagnosis project | coronavirus ({COVID-19}) assessment tool,
  \url{https://www.humandx.org/covid-19/assessment}, (Accessed on 10/21/2020).

\bibitem{NuovoCor51:online}
Nuovo coronavirus covid-19 - paginemediche,
  \url{https://www.paginemediche.it/coronavirus}, (Accessed on 10/21/2020).

\bibitem{httpsasi60:online}
M.~d.~S. Gobierno~de España, Asistenciacovid19,
  \url{https://asistencia.covid19.gob.es/}, (Accessed on 10/21/2020).

\bibitem{Infosfur89:online}
Infos für wirtschaft - infos zum coronavirus,
  \url{https://coronavirus.wien.gv.at/site/wirtschaft/}, (Accessed on
  10/21/2020).

\bibitem{Hippokra35:online}
Hippokrates it gmbh, \url{https://hippokrates-it.de/corona/}, (Accessed on
  10/21/2020).

\bibitem{intensivecare:online}
Suspected {COVID-19} pneumonia diagnosis aid system,
  \url{https://intensivecare.shinyapps.io/COVID19/}, (Accessed on 10/21/2020).

\bibitem{herokuapp:online}
Ml-based {COVID-19} test from routine blood test,
  \url{https://covid19-blood-ml.herokuapp.com/}, (Accessed on 10/21/2020).

\bibitem{Feng2020.03.19.20039099}
C.~Feng, Z.~Huang, L.~Wang, X.~Chen, Y.~Zhai, F.~Zhu, H.~Chen, Y.~Wang, X.~Su,
  S.~Huang, L.~Tian, W.~Zhu, W.~Sun, L.~Zhang, Q.~Han, J.~Zhang, F.~Pan,
  L.~Chen, Z.~Zhu, H.~Xiao, Y.~Liu, G.~Liu, W.~Chen, T.~Li,
  \href{https://www.medrxiv.org/content/early/2020/03/20/2020.03.19.20039099}{A
  novel triage tool of artificial intelligence assisted diagnosis aid system
  for suspected {COVID-19} pneumonia in fever clinics}, medRxiv\href
  {http://dx.doi.org/10.1101/2020.03.19.20039099}
  {\path{doi:10.1101/2020.03.19.20039099}}.
\newline\urlprefix\url{https://www.medrxiv.org/content/early/2020/03/20/2020.03.19.20039099}

\bibitem{Brinati2020.04.22.20075143}
D.~Brinati, A.~Campagner, D.~Ferrari, M.~Locatelli, G.~Banfi, F.~Cabitza,
  \href{https://www.medrxiv.org/content/early/2020/04/25/2020.04.22.20075143}{Detection
  of {COVID-19} infection from routine blood exams with machine learning: a
  feasibility study}, medRxiv\href
  {http://arxiv.org/abs/https://www.medrxiv.org/content/early/2020/04/25/2020.04.22.20075143.full.pdf}
  {\path{arXiv:https://www.medrxiv.org/content/early/2020/04/25/2020.04.22.20075143.full.pdf}},
  \href {http://dx.doi.org/10.1101/2020.04.22.20075143}
  {\path{doi:10.1101/2020.04.22.20075143}}.
\newline\urlprefix\url{https://www.medrxiv.org/content/early/2020/04/25/2020.04.22.20075143}

\bibitem{6772729}
R.~W. {Hamming}, Error detecting and error correcting codes, The Bell System
  Technical Journal 29~(2) (1950) 147--160.

\bibitem{doi:10.1177/001316446602600402}
L.~L. McQuitty, \href{https://doi.org/10.1177/001316446602600402}{Similarity
  analysis by reciprocal pairs for discrete and continuous data}, Educational
  and Psychological Measurement 26~(4) (1966) 825--831.
\newblock \href
  {http://arxiv.org/abs/https://doi.org/10.1177/001316446602600402}
  {\path{arXiv:https://doi.org/10.1177/001316446602600402}}, \href
  {http://dx.doi.org/10.1177/001316446602600402}
  {\path{doi:10.1177/001316446602600402}}.
\newline\urlprefix\url{https://doi.org/10.1177/001316446602600402}

\bibitem{mccullagh1989generalized}
P.~McCullagh, J.~Nelder,
  \href{http://books.google.com/books?id=h9kFH2\_FfBkC}{Generalized Linear
  Models, Second Edition}, Chapman and Hall/CRC Monographs on Statistics and
  Applied Probability Series, Chapman \& Hall, 1989.
\newline\urlprefix\url{http://books.google.com/books?id=h9kFH2\_FfBkC}

\bibitem{10.1145/2939672.2939785}
T.~Chen, C.~Guestrin, \href{https://doi.org/10.1145/2939672.2939785}{{XGBoost}:
  A scalable tree boosting system}, in: Proceedings of the 22nd ACM SIGKDD
  International Conference on Knowledge Discovery and Data Mining, KDD '16,
  Association for Computing Machinery, New York, NY, USA, 2016, p. 785–794.
\newblock \href {http://dx.doi.org/10.1145/2939672.2939785}
  {\path{doi:10.1145/2939672.2939785}}.
\newline\urlprefix\url{https://doi.org/10.1145/2939672.2939785}

\bibitem{Renv}
{R Core Team}, \href{https://www.R-project.org}{R: A Language and Environment
  for Statistical Computing}, R Foundation for Statistical Computing, Vienna,
  Austria (2020).
\newline\urlprefix\url{https://www.R-project.org}

\bibitem{npv_ppv__reas}
R.~Trevethan,
  \href{https://www.frontiersin.org/article/10.3389/fpubh.2017.00307}{Sensitivity,
  specificity, and predictive values: Foundations, pliabilities, and pitfalls
  in research and practice}, Frontiers in Public Health 5 (2017) 307.
\newblock \href {http://dx.doi.org/10.3389/fpubh.2017.00307}
  {\path{doi:10.3389/fpubh.2017.00307}}.
\newline\urlprefix\url{https://www.frontiersin.org/article/10.3389/fpubh.2017.00307}

\bibitem{cramer1946mathematical}
H.~Cramér,
  \href{https://www.degruyter.com/princetonup/view/book/9781400883868/10.1515/9781400883868-fm.xml}{Mathematical
  Methods of Statistics}, Princeton University Press, Princeton, 31 Dec. 1946.
\newblock \href {http://dx.doi.org/https://doi.org/10.1515/9781400883868-fm}
  {\path{doi:https://doi.org/10.1515/9781400883868-fm}}.
\newline\urlprefix\url{https://www.degruyter.com/princetonup/view/book/9781400883868/10.1515/9781400883868-fm.xml}

\bibitem{cohen2013statistical}
J.~Cohen, Statistical power analysis for the behavioral sciences, Academic
  press, 2013.

\bibitem{wagenmakers2007practical}
E.-J. Wagenmakers, A practical solution to the pervasive problems ofp values,
  Psychonomic bulletin \& review 14~(5) (2007) 779--804.

\bibitem{jeffreys1998theory}
H.~Jeffreys, The theory of probability, OUP Oxford, 1998.

\bibitem{scr_test}
L.~D. Maxim, R.~Niebo, M.~J. Utell,
  \href{https://doi.org/10.3109/08958378.2014.955932}{Screening tests: a review
  with examples}, Inhalation Toxicology 26~(13) (2014) 811--828, pMID:
  25264934.
\newblock \href
  {http://arxiv.org/abs/https://doi.org/10.3109/08958378.2014.955932}
  {\path{arXiv:https://doi.org/10.3109/08958378.2014.955932}}, \href
  {http://dx.doi.org/10.3109/08958378.2014.955932}
  {\path{doi:10.3109/08958378.2014.955932}}.
\newline\urlprefix\url{https://doi.org/10.3109/08958378.2014.955932}

\bibitem{autoxgboost}
J.~Thomas, S.~Coors, B.~Bischl, Automatic gradient boosting, in: International
  Workshop on Automatic Machine Learning at ICML, 2018.

\bibitem{Therneau2109rpart}
T.~Therneau, B.~Atkinson, B.~Ripley, Package ‘rpart’,
  https://cran.r-project.org/web/packages/rpart/rpart.pdf, accessed 2020-10-31
  (2019).

\bibitem{Meng_COVID_2020}
X.~Meng, Y.~Deng, Z.~Dai, Z.~Meng, {COVID}-19 and anosmia: A review based on
  up-to-date knowledge., American journal of otolaryngology 41 (2020) 102581.

\bibitem{Pierron_Smell_2020}
D.~Pierron, V.~Pereda-Loth, M.~Mantel, M.~Moranges, E.~Bignon, O.~Alva,
  J.~Kabous, M.~Heiske, J.~Pacalon, R.~David, C.~Dinnella, S.~Spinelli,
  E.~Monteleone, M.~C. Farruggia, K.~W. Cooper, E.~A. Sell, T.~Thomas-Danguin,
  A.~J. Bakke, V.~Parma, J.~E. Hayes, T.~Letellier, C.~Ferdenzi,
  J.~Golebiowski, M.~Bensafi, Smell and taste changes are early indicators of
  the {COVID-19} pandemic and political decision effectiveness., Nature
  communications 11 (2020) 5152.

\end{thebibliography}

\end{document}